\documentclass[final]{cvpr}

\usepackage{times}
\usepackage{epsfig}
\usepackage{graphicx}
\usepackage{amsmath}
\usepackage{amssymb}
\usepackage{capt-of}
\usepackage{comment}
\usepackage[pagebackref=true,breaklinks=true,colorlinks,bookmarks=false]{hyperref}

\def\cvprPaperID{2191} % *** Enter the CVPR Paper ID here
\def\confYear{CVPR 2021}
\usepackage{multirow}
\usepackage[table,xcdraw]{xcolor}

\newcommand{\ym}[1]{}
\newcommand{\yasu}[1]{}
\newcommand{\rzc}[1]{}
\newcommand{\rz}[1]{}

\newcommand{\mysubsubsection}[1]{\noindent {\bf #1}:}
\newcommand{\mysubsubsubsection}[1]{\noindent {\bf #1}:}
\newcommand{\nickname}{Roof-GAN}

\begin{document}

%%%%%%%%% TITLE
\title{Roof-GAN: Learning to Generate Roof Geometry and Relations\\ for Residential Houses}

\author{Yiming Qian \qquad Hao Zhang \qquad Yasutaka Furukawa\\
Simon Fraser University\\
{\tt\small \{yimingq, haoz, furukawa\}@sfu.ca}
}

\twocolumn[{
\maketitle
\vspace{-2em}
\centerline{
\includegraphics[width=0.99\linewidth]{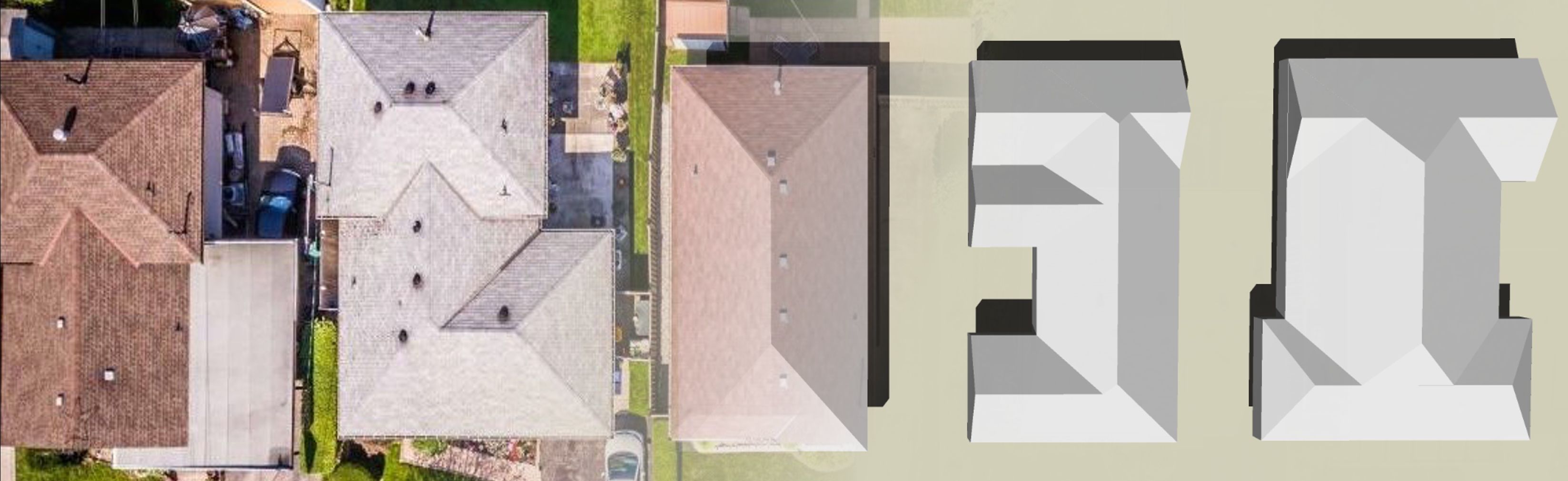}
}
\captionof{figure}{The paper proposes a novel generative model, producing realistic residential roof models. An aerial shot of the real houses are at the left. Our generated samples are at the right. In the middle, we overlay a real house and our generation with similar structure.}
\label{fig:teaser}
\vspace{1em}
}]

% \pagestyle{empty}  % no page number for the second and the later pages
% \thispagestyle{empty} %

%%%%%%%%% ABSTRACT

%%%%%%%%% BODY TEXT
\begin{abstract}
This paper presents Roof-GAN, a novel generative adversarial network that generates structured geometry of residential roof structures as a set of roof primitives and their relationships. Given the number of primitives, the generator produces a structured roof model as a graph, which consists of 1) primitive geometry as raster images at each node, encoding facet segmentation and angles; 2) inter-primitive colinear/coplanar relationships at each edge; and 3) primitive geometry in a vector format at each node, generated by a novel differentiable vectorizer while enforcing the relationships. The discriminator is trained to assess the primitive raster geometry, the primitive relationships, and the primitive vector geometry in a fully end-to-end architecture. Qualitative and quantitative evaluations demonstrate the effectiveness of our approach in generating diverse and realistic roof models over the competing methods with a novel metric proposed in this paper for the task of structured geometry generation.
Code and data are available at \small{\url{https://github.com/yi-ming-qian/roofgan}}.
% We will share our code and data.
\end{abstract}
\section{Introduction}
Residential roof structure exhibits intricate structural details and regularities. 
%
%Careful observation reveals that a complex polygonal surface structure emerges from a combination of a few number of primitive shapes under 
An observation reveals that a complex polygonal surface structure emerges from a careful combination of a few primitive shapes under incident geometric relationships such as colinearity or coplanarity.
%A novel differentiable geometry alignment module further generates adjusted geometry images that satisfy the relationships.
%A few number of primitive shapes yield an interesting combination of complex polygonal surface structure
%A few number of primitive shapes yield interesting 
%The roof structure is a combination of a few primitives. 
%. Careful examination reveals that the roof structure is often a set of a few primitives, yielding complex combination of flat or slanted surfaces.
%Architectural components such as walls, windows, floors are mostly flat, while roofs are a distinct element due to large structural variations. The shape of a roof differs greatly from region to region, varying from a flat surface to a complex combination of slanted faces. Nevertheless, a man-made roof is full of intrinsic structural regularities. For instance, spatial colinearity constrains region layout, and surface coplanrity induces face orientation. 
Automated generation of high quality residential house models and any man-made structures beyond would have tremendous impact on broader disciplines such as construction, manufacturing, urban planning, and visual effects.
%Modeling these relationships is important in making high-quality roofs, which has numerous applications in urban planning, construction and entertainment.\ym{citation?}

With the emergence of deep learning, automated generation of CAD-style 3D models has seen a breakthrough. Early methods
%3D generative methods have seen breakthrough~\cite{moschoglou20193dfacegan,chen2019learning}. For CAD-style 3D model generation, early methods 
focus on generating geometry without their incident relationships 
%by producing a set of geometric parameters
%
as a part assembly~\cite{huang2015analysis,nash2017shape,wu2020pq}.
Auto-encoder based methods learn to generate both geometry and relationships in a form of a binary tree~\cite{li2017grass}, a hierarchical N-ary tree~\cite{mo2019structurenet}, or deformable mesh models~\cite{gao2019sdm}. 
These techniques use fully connected layers with 1D feature vectors to produce CAD geometries in a vector format.

%learn inductive bias in the parameter 
%exploit inductive bias in the algebraic parameter representation of CAD geometry.

The paper takes the CAD-geometry generation research to the next level, while making the following distinctions from the existing methods: 
1) Adversarial training is the foundation of our architecture, providing real generative power over auto-encoder based methods; 2) Convolution with raster-geometry representation enables effective spatial part arrangements and incident relationship generation; and 3) A novel differentiable vectorization module generates vector-geometry in an end-to-end architecture.
% to vectorize raster-geometry with the generated incident relationships.
%bridges the gap between the raster representation and the vector output.
%utilizes generated incident relationships in an end-to-end architecture.

%ncident relationship generation with differentiable geometry alignment allows to learn vectorization

% where the relationship generation and differentiable geometry alignment module makes the vectorization post-processing simple.

% \cite{li2017grass,mo2019structurenet,gao2019sdm}

%realistic human faces, 3D chairs \cite{chen2019learning}, and even fine-grained 3D parts \cite{li2017grass,mo2019structurenet}. However, these methods focus on geometry and do not generate their relationships~\cite{chen2019learning,wu2020pq}, not suitable for the generation of architecture where part incident relations are important. Auto-encoder based approaches take geometry and relations 

%or use autoencoders to reconstruct relations, where subsequent generation suffers from reconstruction quality \cite{li2017grass,mo2019structurenet}. Hence, directly applying them for roof generation cannot produce compelling results for real-world applications, considering the rich relational structures. Moreover, relationship reasoning from popular geometry representations such as mesh and point cloud is known to be a hard problem. Thus, there is a need for better representations that could facilitate architectural relationship enforcement.

Concretely, we propose a novel generative adversarial network, dubbed Roof-GAN.
Given the number of primitives, the generator
%initializes a graph of noise vectors and 
produces a structured geometry model as a graph. A node contains primitive geometry information as a 4-channel image (\ie, roof facet segmentation and roof angles). An edge contains incident primitive relationships as one-hot vectors (\ie, colinearity of footprint boundaries and parallelism of facets, where colinear and parallel implies coplanar).
A node also contains geometry information in a vector format from a differentiable vectorizer, which is further capable of enforcing incident relationships.
%in the vectorization process.
%
Roof-GAN employs two discriminators, one for assessing holistic geometry composition and the other for examining relationship labels together with geometry. 

We have evaluated the proposed approach against the current state-of-the-art, while creating a new database of CAD-style roof geometry with incident relationships, consisting of $502$ residential houses.
%by exploiting aerial LiDAR data over England~\cite{LiDAR}.
%database of ground-truth (GT) roof structures from residential houses in England, by manually annotating roof regions and roof types, and obtaining GT roof face orientations from associated depthmaps. Our 
Qualitative and quantitative evaluations demonstrate the effectiveness of \nickname~against competing methods in generating diverse and realistic set of roof models with a novel metric proposed in this paper for the task of structured geometry generation. Code and data are available at \url{https://github.com/yi-ming-qian/roofgan}.
\section{Related Work}

We review related techniques in three domains: architectural reconstruction, generative models for architectural structures, and assembly-based modeling.

\mysubsubsection{Architectural reconstruction}
Reconstruction of architectural elements such as lines, planes, room layouts, %floorplans, 
and 3D buildings has a long history in vision research.
Traditional methods are either rule-based, e.g., built on shape grammars~\cite{dick2002bayesian,lin2013semantic,huang2013generative}, or use optimization with ad-hoc objectives, typically requiring depth or multiple view information to infer planes~\cite{furukawa2009manhattan}, room layouts~\cite{silberman2012indoor}, or CAD-quality objects~\cite{nan2017polyfit}.
%hand-crafted rules and typically require depths or multiple views~\cite{cabral2014piecewise,furukawa2009manhattan,nan2017polyfit,silberman2012indoor,lin2019floorplan} as input. 
%For full 3D buildings, shape grammars are employed to constraint the model space~\cite{dick2002bayesian,lin2013semantic,huang2013generative}.
%
With the surge of deep neural networks (DNNs), data-driven methods enable single-image reconstruction of depthmaps~\cite{liu2019planercnn,qian2020learning}, room layouts~\cite{zhou2019learning}, or wire-frame building models~\cite{zou2018layoutnet}. Shape grammars are also utilized for roof reconstruction by DNNs, which classify grammar branches and estimate geometric parameters~\cite{zeng2018neural}. Recently, Conv-MPN~\cite{zhang2020conv} learns to reconstruct vector-graphics building models, without shape grammars, by using a relational neural architecture, which is the backbone of our network.
In contrast to reconstruction, our RoofGAN is a fully generative model, striving for plausibility and diversity.

%For full 3D buildings, reconstruction algorithms utilize shape grammars to constraint the model space via hand-coded algorithms 
%cedural reconstruction relies on shape grammars for buildings, where the grammar-rules are parsed by either 
%hand-coded algorithms \cite{dick2002bayesian,lin2013semantic,huang2013generative} or DNNs \cite{zeng2018neural,nishida2018procedural} to classify grammar types and estimate geometrical parameters. This paper instead targets at automated generation of residential house roofs.

\mysubsubsection{Generative models for architecture} Procedural models with hand-crafted rules can provide production-level solutions to virtual building generation~\cite{muller2006procedural}. %, where hand-crafted shape grammars are applied in an iterative fashion.
Various techniques have been leveraged to improve the generation quality such as Markov Chain Monte Carlo \cite{talton2011metropolis}, discrete optimization \cite{hendrikx2013procedural} and, probabilistic graphical model \cite{merrell2010computer}.

Inspired by the success in deep image generation, several deep generative networks have been proposed for 2D or 3D structured data~\cite{sid20203dgen}. Recursive neural networks have been applied to learn object placements and relations for indoor scene generation \cite{li2019grains}, while graph CNNs have been trained to generate room layouts~\cite{wang2019planit} or floorplans \cite{Graph2Plan20,wu2019data} by providing building outlines as input.
%
%Using a three-step DNN, ArchiGAN \cite{chaillou2019ai+} %couples the two problems by 
%jointly generates building floorprints, floorplans, and furniture arrangements.} 
Closer to our work, House-GAN \cite{nauata2020house} employs a generative adverserial network for floorplan generation, while requiring room adjacency relations as input. \nickname~also follows adverserial training, but does not input building outlines or adjacency relations. Instead, the relations are generated by the network.
\\
\mysubsubsection{Assembly-based modeling.} Many methods have been proposed for 3D shape generation using different representations: voxels~\cite{wu2016learning,girdhar2016learning}, point clouds~\cite{achlioptas2018learning,fan2017point}, meshes~\cite{chen2020bsp}, and implicit functions~\cite{chen2019learning,park2019deepsdf}. 
Generative models for 3D shape structures~\cite{xu2016data,sid20203dgen} are typically trained to learn models of part variations and assembly.
% have been propelled by
%These methods mainly focus on intra-part geometry due to their holistic representation. Since 
%the release of structured shape repositories such as PartNet \cite{mo2019partnet}. 
%3D part geneartion has become an active research topic. 
%In addition to per-part geometry, researchers started to acknowledge the importance of modeling inter-part relationship (\eg adjacency and symmetry) for generating realistic 3D parts. 
Most representative works resort to part and structure autoencoders, that are built on different shape structure representations, including hierarchical trees in GRASS~\cite{li2017grass}, graphs in StructureNet~\cite{mo2019structurenet}, part assembly sequences in PQ-Net~\cite{wu2020pq}, and more general graphs in other works~\cite{schor2019componet,dubrovina2019factor,gao2019sdm}. In contrast, our network is designed with generative capability and diversity in mind, relying on adversarial training which is shown to be effective even with a relative small training set.

%both of which represent 3D parts with a hierarchical tree and use recursive neural network-based autoencoder to encode both geometry and relationships. 
\section{Structured Roof Geometry Representation}
\label{sec:representation}

%The roof geometry is represented as a graph, where a node contains primitive geometry as a raster image and an edge contains their incident relationships

%t easy for convolutional message passing networks in the generator and the discriminators to reason spatial arrangements of part geometry with their incident relationships; and 2) enabling differentiable vectorization in an end-to-end architecture.
%the first contribution of the paper, which makes it easy for convolutonal message passing network to reason spatial arrangement of part geometry, and enables diffentiable vectorization in an end-to-end architecture.
%
%raster encoding, which 1) helps convolutional message passing architecture to generate realistic arrangements; and 2) enables differentiable vectorizer to produce vector-geometry parameters end-to-end.
%
A combination of a few number of rectangular roof primitives explains the majority of residential roof structure. We represent a roof as a graph, where a node encodes primitive geometry information and an edge encodes pairwise incident relationships.
%, in particular, colinearity of footprint boundaries and coplanarity of facets.
Structured geometry representation is at the heart of this research, enabling 1) effective spatial reasoning of primitive arrangement; and 2) differentiable vectorization in an end-to-end architecture.

\begin{figure}
    \centering
    \includegraphics[width=\linewidth]{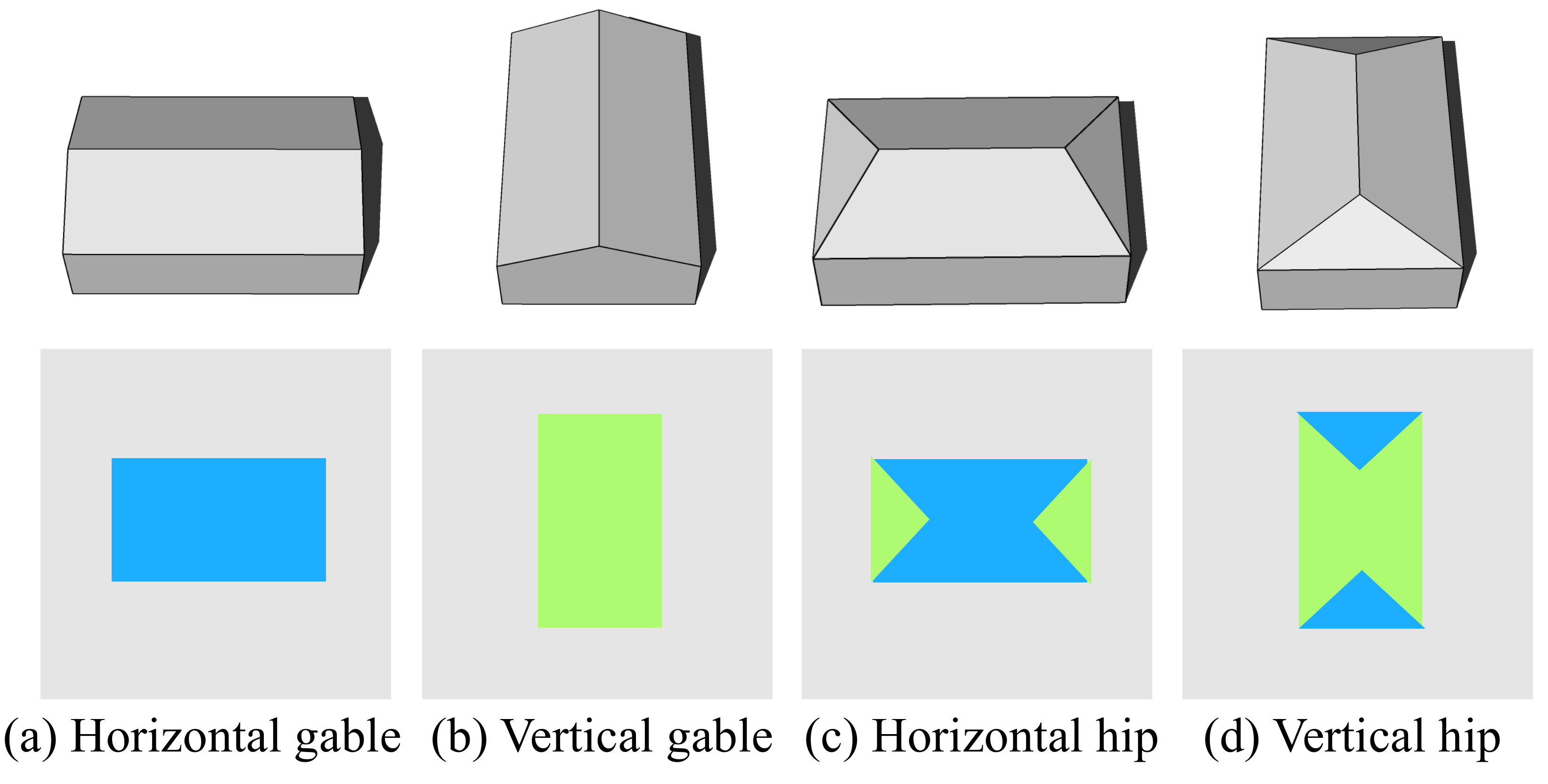}
    \caption{
    %\yasu{The color choice seems to be ugly overall. Pure blue or pure green is often a bad color to fully fill-in a region (bottom). Can we use more beautiful/pastel colors? Use the ones from the figure screen shot I pasted in slack. The top rendering is also weird. Can we use say Sketchup to render a nice 3D model maybe with transparency or anything to make it look beautiful? Maybe there should be a ground place visible in the top row.} 
    %
    Our model handles four types of roof primitives, whose sample 3D models are shown at the top.
    %The top shows sample 3D mesh models for the four different roof primitive types.
    %
    The bottom shows a part of our raster geometry representation, where a 3-dimensional one-hot encoding represents the roof facet type per pixel: top/bottom facing ({\color{blue}blue}), left/right facing ({\color{green}green}), or background (grey). The representation also has a roof-angle image, which stores the angle between the facet normal and the inverse gravity per pixel.
    %
    %Our 2D parameterization for a roof region. The top row shows example 3D mesh models for each roof type and the bottom row shows the corresponding 2D roof-face segmentation mask. In particular, we label the roof structure with three classes: background (grey), top and bottom faces ({\color{blue}blue}), left and right faces ({\color{green}green}). For example, the horizontal gable contains top and bottom faces only, so its mask does not have green colors.
    }
    \label{fig:representation}
\end{figure}

\mysubsubsection{Geometry representation at nodes}
%A roof can be segmented into multiple overlapped regions, where each region has an individual top covering and is typically associated with a room of the house.
We assume that each primitive is an axis-aligned rectangle. The top covering is either horizontal-gable, vertical-gable, horizontal-hip, or vertical-hip (See Fig.~\ref{fig:representation}).
%A standard form is to prepare different representations for different roof types~\cite{zeng2018neural}, which have different topology.
%
%
%We represent a roof as a graph, where the nodes encode region geometry and the edges encode inter-region relationships. In particular, we assume each region is an axis-aligned rectangle and its top covering is categorized as: horizontal gable, vertical gable, horizontal hip or vertical hip. 
%because which is natural because a hip roof and a gable roof have different topology.
%has more degrees of freedom than a gable roof, for example.
%
%could be using separate representations for geometrical parameters and structural semantics (\eg, 1D vectors for region bounding boxes and 4 labels for roof types). In contrast, we propose a new 2D representation that integrates geometry and semantics, as shown in Fig. \ref{fig:representation}. 
A 4-channel image is our representation, where the first three channels constitute one-hot encoding over the three facet \emph{orientations}: 1) left/right facing; 2) top/bottom facing; and 3) background. The fourth channel is the facet \emph{angle} (\ie, the angle between the normal and the inverse gravity), which is set to 0 for background.

Differentiable vectorization (Sect.~\ref{sec:vectorization}) converts a 4-channel image into a 6d vector.
%in our architecture.
% As shown in Sect.~\ref{sec:vectorization}, a differentiable vectorization module converts the 4-channel image into the vector format.
The six numbers are the four boundary coordinates of the rectangle and the two roof angles (one for left/right facets and one for top/bottom facets). Note that we are further assuming that
%, which are nonetheless true to the majority of residential houses: 
1) Primitives are symmetric, where the angles of the left/right (resp. the top/bottom) facets are equal. 2) The height of the wall is a constant (not generated). 3) When primitives overlap, we keep the highest facet and ignore invisible portions.

%We assume that each primitive is symmetric. For example, the angles of the top and the bottom facets are equivalent for a horizontal gable, and triangularf facets in the hip roofs are 

%Our representation for each node\footnote{The terms ``node'' and ``region'' will be interchangeably used in the following sections.} are two 2D images: one for roof structure segmentation and another for roof face orientation. Specifically, for the rasterized image of a region observed from the gravity direction,
%we label each pixel by three categories: (1) top and bottom faces, (2) left and right faces, (3) background, resulting in a 3-channel image by one-hot encoding. Note that here we assign the same label to the symmetric roof faces, which naturally allows to generate symmetric structures. Similarly, we create another image by assigning each pixel with the orientation of the corresponding roof face, which is defined as the angular value between the roof normal and the inverse gravity direction. We set $0$ to background pixels.

\mysubsubsection{Relationship representation at edges}
For each pair of primitives, 6d
%\ym{here it should be 6d, instead of 8d. The facets are symmetric for top/bottom, left/right} 
one-hot encoding represents colinearity of rectangular boundaries and parallelism of polygonal facets.
Concretely, the first dimension encodes if the left rectangular boundaries are colinear. The next three dimensions are for the colinearity of the right/top/bottom boundaries. We do not model a colinearity of the left boundary and the right boundary, which rarely happens in our data because primitives overlap instead. The remaining two dimensions encode the parallelism of the facets of the primitive pair (one for the left/right facets and one for top/bottom facets), where facets are coplanar when they are parallel and the corresponding boundaries are colinear.

%We consider two types of pairwise inter-node relationships.\\
%\noindent $\bullet$ {\bf Colinearity}: Are two region boundaries colinear? \\
%\noindent $\bullet$ {\bf Coplanrity}: Are two roof faces coplanar?\\
%\noindent
%It is prevalent that multiple regions are sharing the same wall of a house. For each region pair, we assume four colinearity types: colinear top-top boundaries, colinear bottom-bottom boundaries, colinear left-left boundaries, and colinear right-right boundaries. Likewise, we assume four corresponding coplanar faces. 

\section{\nickname}
\begin{figure*}[tb]
    \centering
    \includegraphics[width=\textwidth]{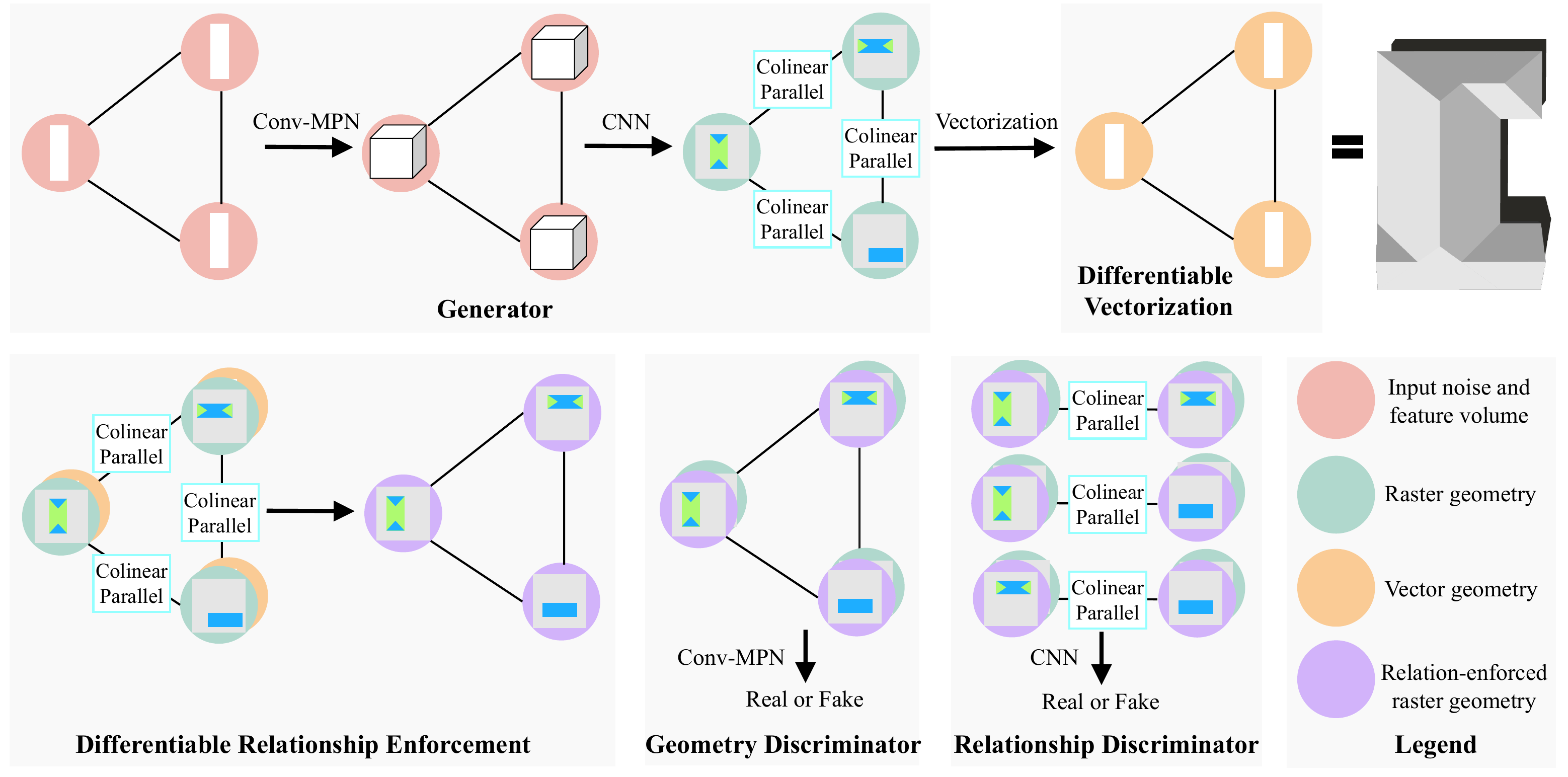}
    \caption{{\nickname} architecture. The generator takes a graph of noise vectors as input and utilize Conv-MPN \cite{zhang2020conv} to obtain 3D feature volumnes. Several convoluational layers are used to output per-node raster-geometry (here we draw orientation images only for illustration) and pairwise relationships. A differentiable vectorization module converts the raster-geometry to the vector-geometry, followed by a differentiable relationship enforcement. The geometry discriminator takes the raster-geometry before and after the relationship-enforcement for all the nodes. The relationship discriminator acts on a pair of nodes while also taking the relationship information as the input.
    %further takes the relationship information for each pair of nodes. 
    %takes a pair of     outputs a scalar regarding geometry realism, while the relationship discriminator outputs a scalar regrading relationship realism for each pair of nodes.
    }
    \label{fig:network}
\end{figure*}

Relational GAN for floorplan generation is our backbone~\cite{nauata2020house}.
Given the number of primitives, Roof-GAN initializes a complete graph of noise vectors and generates a roof geometry (See Fig.~\ref{fig:network}).
We focus on the differences from the prior work,
%(\ie, relationship generation, differentiable vectorization, and relationship discrimination),
%(generator, differentiable vectorization, and discriminator),
where the architectural details are in the supplementary document and
%, while we refer the complete architectural details to the supplementary document
%Three differences in the architecture make our system unique: generator, differentiable vectorization, and discriminator 
Section~\ref{sec:results} explains how we pick the number of primitives at training and testing.

%Our method belongs to the family of generative adversarial networks. Suppose the number of roof regions $K$ is given as a prior, we employ a graph neural network-based generator to generate per-region geometry and inter-region relationships, which are passed to two separate discriminators, respectively. During training, we set $K$ using the GT number of regions in the same batch, whereas $K$ is randomly chosen from $[2,5]$ at test time.

\subsection{Generator}
%In the prior work~\cite{nauata2020house}, the generator takes a graph of components whose nodes encode room types and edges encode their relationships.
%Roof-GAN generates geometry at nodes and relationships at edges, where the only input is the number of primitives. 
%node types and relationships as opposed to receiving as an input. 
%Therefore, the generator is a complete relational graph.
%nput is a complete graph without any additional information.
%Note that during training, $K$ is chosen from $[2, 5]$ at random per batch, with the same proportions as in the ground-truth samples. At test time, $K$ is chosen from $[2, 5]$ with a uniform random probability.
%\yasu{Cross-validation?}

%\mysubsubsection{Input layer}
%Feature extraction}
%We form a \textit{complete} graph where each node encodes a rectangular region.
Each node is initialized with a 128d noise vector sampled from the normal distribution. Note that there are no one-hot node-type vectors as in the prior work~\cite{nauata2020house}, because node/edge properties are to be generated instead of given in our work.
%no properties are associated with nodes/edges in our case.
%and  (w/o a one-hot vector encoding node properties as in the prior work~\cite{nauata2020house}).
%, without the one-hot vector encoding a node type~\cite{nauata2020house}.
The architecture is the same till the output layers, where
%The rest of the architecture till outputs is the same~\cite{nauata2020house}, which reshapes a feature vector to a 3D tensor, and utilizes convolutional message passing architecture (a variant of graph neural network) to produce a graph of feature volumes per node.
%
%The input to each node is a $128$D noise vector sampled from normal distribution. We then apply a shared linear layer to map it to a $1024$D vector, which is subsequently reshaped to a 3D tensor $\mathbf{H}_i$ of size $16\textrm{-channel}\times8\textrm{-row}\times8\textrm{-column}$.
%
%To encode the graph with initial node features $\{\mathbf{F}_i^0=\mathbf{H}_i|1\leq i\leq K\}$, we follow the Conv-MPN \cite{zhang2020conv,nauata2020house} architecture to perform two iterations of message passing for feature update:
%\begin{equation}
    %\mathbf{F}_i^t = \textrm{Upsample}^t\left(\textrm{CNN}^{t}\left(\mathbf{F}_i^{t-1}; \underset{1\leq i\leq K, j\neq i}{\textrm{MaxPool}} \mathbf{F}_j^{t-1}\right)\right),
%    \label{eq:convmpn}
%\end{equation}
%where $t\in\{1,2\}$ is the iteration index. More specifically, for feature $\mathbf{F}_i^{t-1}$ at each node, we firstly concatenate it with the max-pooled features over other nodes. Then a 3-layer $\textrm{CNN}^{t}$ is applied to update features, followed by a transposed convolutional layer for $2\times$ upsampling. Please see the detailed specification in the supplemental materials. Finally, we obtain the updated features $\{\mathbf{F}_i^2|1\leq i\leq K\}$ (each has the size of $16\times32\times32$).
%\mysubsubsection{Geometry outputs}
the last feature volume per node has the dimension of $16\times 32\times 32$.
Two 3-layer CNNs are used to produce a $3\times 32\times 32$ facet orientation image and a $1\times 32\times 32$ facet angle image.
%Given $\mathbf{F}_i^2$, two 3-layer CNNs are used to predict two images: a $3\times32\times32$ face segmentation mask denoted as $\mathbf{S}_i$, and a $1\times32\times32$ face orientation map denoted as $\mathbf{N}_i$, respectively. 
A softmax function is applied for the orientation image and a sigmoid function is applied for the angle image as the cosine value of the angle.
%
%We conduct pairwise relationship generation. 
For each node pair, we concatenate node features into a
%$(\mathbf{F}_i^2; \mathbf{F}_j^2)$, which becomes the corresponding edge 
$32\times32\times32$ volume, and apply a 5-layer CNN to downsample it into a $512\times1\times1$ vector.  Finally, we use a linear layer and a sigmoid function to convert to a 6d relationship vector. Note that the node feature concatenation order is arbitrary.
%depends on the node order, which is set to arbitrary.
%$\mathbf{r}_{i,j}$, where the first four values correspond to four colinearity types and the other four values correspond to four coplanrity types.

\subsection{Differentiable vectorization} \label{sec:vectorization}
%The architecture utilizes raster geometry representation for powerful spatial reasoning through a convolutional message passing network (Conv-MPN).
A novel differentiable vectorization converts the raster geometry into the vector format, serving two purposes. First, this eradicates the post-processing heuristics for vectorization. Second, raster-geometry may not follow the relationships.
%, where the discriminator employs Conv-MPN with the raster representation as input.
%, where it is easy to enforce the relationships by adjusting geometry in the vector space.
The vector parameterization allows us to generate relationship-enforced raster-geometry, which can be passed to the discriminator for assessment. 
There are two vectorization modules, one for rectangle boundary coordinates and the other for roof angles/type. We here describe the former and refer the latter to the supplementary document, which is rather straightforward with a sequence of reasonable heuristics.
%We here describe two of such modules and refer the details of the facet angle regression and the primitive type classification to the supplementary document.
Note that both modules are fixed algebraic machinery without learnable weights. 
Zhang \etal proposed a neat vectorization trick for corner detection, which takes a corner probability image from CNN and computes their weighted mean coordinate~\cite{zhang2018unsupervised}. We extend the idea to rectangle boundary coordinate vectorization.
%We extend a vectorization trick, which was originally proposed for a corner coordinate regression by taking a corner probability image and computing the weighted mean coordinate based on the probability~\cite{zhang2018unsupervised}.
%
Given a facet orientation image (\ie, a probability image over left/right, top/bottom, or background classes), we obtain a primitive mask probability by one minus the background probability. Let us use the left-boundary coordinate as an example. We compute the x-derivative of the mask by finite difference, apply ReLU to keep only non-negative responses, then take the weighted mean coordinate~\cite{zhang2018unsupervised}. The non-negative responses should concentrate on the left boundary of the mask and this algebraic formula computes the left boundary coordinate. Exactly the same algebraic operations apply to the other three boundaries.

\subsection{Differentiable relationship enforcement}
Given rectangle boundary coordinates,
%we obtain colinearity enforced boundary coordinates 
%we obtain relationship-enforced raster geometry images 
we 1) enforce their colinearity relationships and obtain adjusted boundary coordinates probabilistically; 2) solve for the non-uniform scale and translation that maps the original rectangle to the adjusted one; and 3) warp the facet orientation/angle images based on the transformation to obtain relationship-enforced geometry-images. Note that roof angle generation is more stable, and we do not enforce the parallelism relationships with the roof angle image.

%compute refined boundary coordinates by enforcing the colinearity 
%then applying non-uniform scaling and translation to each facet orientation/angle images.
%Note that we do not use facet angles and types in this step,
%which are used only for the generation of vector geometry during testing.

Without loss of generality, the adjusted left coordinate of one primitive is calculated as the weighted average of the left coordinates of all the primitives. The weight is 1 for itself and the colinear relationship probabilities for the others.~\footnote{Low-probability relationship should not influence, and the weight formula is in fact $\mbox{ReLU}(2 p - 1)$, where $p$ is the original probability.}
%Concretely, for each boundary coordinate, say the left coordinate of a reference primitive, we take the weighted average of the estimated left coordinates, where the weight is 1 for the reference primitive itself and is the colinear relationship probabilities for the others.~\footnote{We do not want to influence when the probability is low, and use $\mbox{ReLU}(2 p - 1)$ as the weight formula, where $p$ is the original probability.}
%We use the same formula to compute refined coordinates for the others. 
Given the original and the adjusted coordinates, we compute non-uniform scaling and translation parameters and apply image-warping to the facet orientation/angle images by ``grid\_sample'' built-in function in PyTorch.

%We then snap the boundaries based on the inter-region relationships. Take the top boundary for example, and denote the predicted top boundary colinearity probability as $p_{i,j}$ for each region pair $i$ and $j$. We update $B_i^{top}$ as:
%\begin{equation}
    %\widehat{B}_i^{top} = \frac{\sum_{j=1}^{K} %\textrm{ReLU}(2p_{i,j}-1)B_j^{top}}{\sum_{j=1}^{K} %\textrm{ReLU}(2p_{i,j}-1)},
%\end{equation}
%where we let $p_{i,i}=1$. $\widehat{B}_i^{bottom}, \widehat{B}_i^{left}, \widehat{B}_i^{right}$ are estimated in a similar way.

%Then, an affine transformation between the original $B_i$ and the snapped $\widehat{B}_i$ is estimated by:
%\begin{equation}
%\begin{bmatrix}
%B_i^{left} & B_i^{right} \\
%B_i^{top} & B_i^{bottom}
%\end{bmatrix}=
%\begin{bmatrix}
%s_x & 0 & t_x \\
%0 & s_y & t_x
%\end{bmatrix}
%\begin{bmatrix}
%\widehat{B}_i^{left} & \widehat{B}_i^{right} \\
%\widehat{B}_i^{top} & \widehat{B}_i^{bottom} \\
%1 & 1
%\end{bmatrix},
%\end{equation}
%where $s_x, s_y$ are the scaling factors, and $t_x,t_y$ the translation %factors.

%Finally, given the affine matrix, we warp mask $\mathbf{S}_i$ and orientation map $\mathbf{N}_i$ using bilinear interpolation \cite{jaderberg2015spatial} to obtain the snapped versions $\mathbf{\widehat{S}}_i$ and $\mathbf{\widehat{N}}_i$.

%Note that this module is not used in the relationship-enforcement step and will not influence the discriminator.

%used only at test time for vector-geometry generation but not during training.

\subsection{Discriminators}
Roof-GAN has two discriminators, whose loss functions are added with equal weights.
%Two discriminators are trained to distinguish generated samples from the ground-truth, whose losses are added equally.
%with the equal weight. 
%WGAN-GP \cite{gulrajani2017improved} is used for training.
%and the loss function is linear.\ym{it is confusing by saying such a linear thing. I never saw people write this. You can say ``The discriminator is learned to be a 1-Lipschitz function'', or please refer to supp for detailed training loss equation.} 
%used to perform real/fake judgements for geometry and relationships, respectively. 
%
%\mysubsubsection{Geometry discriminator}
The first discriminator focuses on geometry without relationships. 
%s the same as in the prior work~\cite{nauata2020house} except for the input.
As in the generator, we form a complete relational graph.
At each node, we concatenate the orientation image and the relationship-enforced orientation image into a $6\times 32 \times 32$ volume. A 3-layer CNN converts to a $16\times 32 \times 32$ volume.
The same architecture converts the angle images into a $16\times 32\times 32$ volume. We further concatenate these two volumes to obtain an input to the Conv-MPN architecture, which is the same as the prior work~\cite{nauata2020house}.
%, where edges have no features.
%
For ground-truth samples, the relationship-enforced images are identical to the original.

%The network takes GT and generated segmentation masks $\{\mathbf{S}_i, \mathbf{\widehat{S}}_i | 1\leq i\leq K\}$, and orientation images $\{\mathbf{N}_i, \mathbf{\widehat{N}}_i | 1\leq i\leq K\}$, as input, and outputs a scalar regarding the realism of geometry. For GT samples, the inputs before DS and after DS are identical; Namely, $\mathbf{S}_i=\mathbf{\widehat{S}}_i$ and $\mathbf{N}_i=\mathbf{\widehat{N}}_i$.

%As in generator, we form a \textit{complete} graph and use Conv-MPN \cite{zhang2020conv} for feature extraction. Firstly, for each node $i$, two 3-layer CNNs are exploited to the concatenated masks $(\mathbf{S}_i; \mathbf{\widehat{S}}_i)$ and the concatenated orientation maps  $(\mathbf{N}_i; \mathbf{\widehat{N}}_i)$, respectively, each producing a feature tensor of size $16\times32\times32$.

%Secondly, we concatenate the resulted mask and orientation features, send to a Conv-MPN module as in Eq. \ref{eq:convmpn} but change upsampling to downsampling, yielding a feature tensor of size $32\times8\times8$ after two iterations of message passing. A 3-layer CNN is then used for further downsampling to a 128D feature vector.

%Lastly, we max-pool across all node feature vectors to obtain a global feature vector (128D) and use a single linear layer to predict a scalar $d^{geometry}$ classifying GT and generated geometry.

%\mysubsubsection{Relationship discriminator}
The second discriminator takes a pair of nodes and an edge. 
A 6d relationship vector at the edge is converted to a 4096d vector with a linear layer, and reshaped to a $4\times 32\times 32$ volume. We concatenate with the facet orientation/angle images before and after the relationship enforcement (\ie, eight tensors), then use a 5-layer CNN to down-sample to $128\times 1\times 1$, followed by a linear layer to output a scalar. The average over all node-pairs is the discrimination score. 
%The losses of the two discriminators are added equally.

%The network is implemented in a pairwise fashion. Take pair $(i,j)$ for example, the network's input is a concatenation of nine tensors: $\mathbf{S}_i$, $\mathbf{N}_i$, $\mathbf{\widehat{S}}_i$, $\mathbf{\widehat{N}}_i$, $\mathbf{S}_j$, $\mathbf{N}_j$, $\mathbf{\widehat{S}}_j$, $\mathbf{\widehat{N}}_j$, and a feature tensor mapped from the relationship vector $\mathbf{r}_{i,j}$\footnote{A linear layer is used to convert the 8D vector $\mathbf{r}_{i,j}$ to a 4096D vector, which is reshaped to a $4\times32\times32$ tensor. It is then concatenated with another $8\times32\times32$ tensor expanded from $\mathbf{r}_{i,j}$ by repeating at every pixel location, resulting in a final tensor of size $12\times32\times32$.}. Again, we have $\mathbf{S}_i=\mathbf{\widehat{S}}_i$ and $\mathbf{N}_i=\mathbf{\widehat{N}}_i$ for GT samples. We use a shared 5-layer CNN to downsample the input to a 128D vector, and a following linear layer to output a scalar $d_{i,j}^{relation}$ depicting the relationship realism for each pair $(i,j)$, as shown in Fig. \ref{fig:network}.

%All in all, we sum up the outputs of the two discriminators, getting a final scalar output:
%\begin{equation}
    %d=d^{geometry} + \frac{2}{K(K-1)}\sum_{i=1,j=i+1}^K d_{i,j}^{relation}.
%\end{equation}
\section{Dataset and Metrics}

Zeng \etal introduced a database of residential houses in England, containing height maps and surface normal maps in the aerial view~\cite{zeng2018neural}, which we borrow to construct our database of structured roof geometry with relationships.
%this database and create structured roof geometry  our GT roof geometry and relationships. 
Quantitative evaluation of vector-graphics geometry is a non-trivial task especially for the generative models. We propose a new metric, dubbed Recursive Minimum Matching Distance, which assess the realism and diversity of the generated samples as in the FID metric, while respecting the vector structure of the geometry representation.

%which are constructed from the aerial LiDAR scans provided by UK Environment Agency.

\subsection{Dataset}
Given a surface normal image of a house in a Nadir view,
%Taking an aerial normal image of a house, 
we use the annotation tool ``Colabeler'' \cite{colabeler} to annotate the bounding box of each rectangle primitive and its corresponding roof type. A total of 502 houses are annotated, out of which (152, 256, 68, 26) houses have (2, 3, 4, 5) rectangles, respectively. The colinearity relationships are obtained by simply checking the equality of the annotated bounding box coordinates with an error tolerance of 1 pixel to account for rare human errors.
%with an error tolerance of 1 pixel.
%
Facet angles are calculated from depthmaps by the pre-processing algorithm of the prior work~\cite{zeng2018neural}.
%with a minor modification.
%, we solve an optimization problem to estimate the facet angles based on the depthmaps.
Parallelism is automatically determined by checking the equality of the angles with an error tolerance of $18^{\circ}$. When detecting the colinearity (resp. parallelism), we enforce the relationship by snapping one coordinate (resp. angle) to the other.
%their angles the same by picking one angle arbitrarily.
%
%mention we keep original bbox coordinates and roof angles
%
%Given the bounding box coordinates and facet angles, we rasterize a depthmap and a facet orientation image.
%GT relationship creation is simple. We use bounding boxes to check boundary coliearity with a threshold of 1 pixel, while the optimized roof face angles are used to check coplanarity with a threshold of $18^{\circ}$.

The next step is to merge connected coplanar (i.e., colinear and parallel) rectangular facets into a single polygonal facet.
For each house, we rasterize all the primitives from their parameters into a common facet orientation image (instance-aware), facet angle image, and a height map, while discarding invisible facet information per pixel based on the height values. We use the rasterized images to identify connected coplanar facets and merge them into a single polygon. The data pre-processing algorithms are all heuristics and their details are referred to the supplementary.
%
%.~\footnote{The idea is to optimize all the geometric parameters by minimizing the discrepancy against the input height map. See supplementary for details}
%to estimate the facet angles of all primitives by following \cite{zeng2018neural}.
%
%Given these, for each pixel inside the bounding box of a primitive, we compute height values (two values for gable, four values for hip) using all facets of the primitive, and label that pixel to the facet with the lowest height values. Such a process returns a height map (0 is set to background pixels) and facet segmentation masks (4 for hip, 2 for gable) for each primitive. After merging the top-bottom facet and merging the left-right facet, we obtain the facet orientation and angle images for each primitive, as defined in Sect. \ref{sec:representation}. 
Finally, we perform standard 8x data augmentation by $90^{\circ}$ rotation and mirroring, yielding a total of 4,016 houses.
%We perform standard data augmentations including rotation by $90^{\circ}$ and horizontal/vertical flip, which give us a total of 4016 samples.

\subsection{Metrics}
For quantitative evaluation of generative models for structured geometry, we propose a novel distance metric for a pair of sets of polygonal 3D models, dubbed Recursive Minimum Matching Distance (RMMD), which measures the realism and the diversity of the generation.~\footnote{
%we propose a novel distance metric between two sets of polygonal 3D models.
Since our roof 3D model can be represented as a 2D polygon in a Nadir view, we present the metric in a case of 2D polygons but the formula applies to general 3D models.}
RMMD is recursively defined for a pair ($s_1, s_2$), each of which is a set of polygonal surface models ($\{m_i\}$), each of which is a set of facets ($\{f_i\}$), each of which is a set of vertices ($\{v_i\}$):
\begin{eqnarray*}
\mbox{D}_{S}(s_1, s_2) = \sum_{m_1 \in s_1} \min^{\star}_{m_2 \in s_2} \frac{\mbox{D}_{M}(m_1,m_2) + \mbox{D}_{M}(m_2, m_1)}{2|s_1|},\\
\mbox{D}_{M}(m_1, m_2) = \sum_{f_1 \in m_1} \min^{\star}_{f_2 \in m_2} \frac{\mbox{D}_{F}(f_1, f_2) + \mbox{D}_{F}(f_2, f_1)}{2|m_1|},\\
\mbox{D}_{F}(f_1, f_2) = \sum_{v_1 \in f_1}  \min^{\star}_{v_2 \in f_2} \frac{\mbox{D}_{V}(v_1, v_2) + \mbox{D}_{V}(v_2, v_1)}{2|f_1|},\\
\mbox{D}_{V}(v_1, v_2) = \left|v_1 - v_2\right|.
\end{eqnarray*}

\noindent
At the bottom of the recursion, the distance $\mbox{D}_V(v_1, v_2)$ between two vertices is their Euclidean distance. The distance $\mbox{D}_F(f_1, f_2)$ between two facets is computed as follows. For each vertex in $f_1$, we find the closest vertex from $f_2$ based on the vertex-distance. The average over all the vertices in $f_1$ is the facet-distance, except that we enforce mutual exclusiveness in the matching, that is, each vertex in $f_2$ is matched at most once.
The mutual exclusiveness implies that when $f_1$ has more vertices than $f_2$ (topologically inconsistent), some vertices do not have matches and the largest possible distance (\ie, diagonal of the square image) is used for the average calculation ($\star$ denotes a special $\min$ operation with abuse of notation). In practice, after computing all pairwise vertex distances, we use a greedy algorithm to find the minimum distance matching. The model $D_M$ and the set $D_S$ distances are defined in the same way recursively.

RMMD measures geometric and topological differences into a single number in the unit of the vertex Euclidean distance. Note that each distance function is asymmetric and we make it symmetric inside each summation. At the top level, the number of ground-truth samples (512 augmented from 64) is less than the number of samples in a generation (1000), and an asymmetric distance $D(g_1, g_2)$ is used as our metric, where $g_1$ is the ground-truth set.
RMMD measures the realism and the diversity, because there must be a similar model for every ground-truth sample.
%In our experiments, $\mbox{RMMD}(s_1, s_2)$ computes the distance between a set ($s_1$) of 512 ground-truth samples (augmented from 64) and a set ($s_2$) of 1,000 generated samples by a method as a performance metric, with one preprocessing. 

Lastly, RMMD is sensitive to scale and translation differences. We normalize each model by taking its axis aligned bounding box and apply uniform scaling and translation so that the bounding box is center-aligned and tightly fits inside a square image, which is assumed to be $16m\times 16m$.

We also use a standard FID metric~\cite{heusel2017gans}. Each roof model is represented as a surface normal image where the 3D surface normal vector is used as a RGB value and the background is set to white. Again, 64 ground-truth samples and 1,000 generated samples are used for the metric calculation.

\section{Experiments} \label{sec:results}
We have implemented \nickname~in PyTorch and trained the models with
%thon based on the PyTorch library. 
%We train the generator and the discriminators with 
WGAN-GP~\cite{gulrajani2017improved} and the Adam optimizer. The learning rate is $0.0001$, the batch size is 16,
%are trained, which
%The batch size of 16 is used.
%Our training follows WGAN-gp \cite{gulrajani2017improved}, where 
the number of critics is 1, and the weight of gradient penalty is 10.
The training takes about 17 hours on an NVIDIA GTX 1080 Ti GPU with 11GB of RAM for 200k iterations. 
%The GT images have the resolution of $64\times64$, which are downsampled by half for faster training. At test time, we generate images 

\subsection{Competing methods}

We have compared against three competing methods and two \nickname~variants for an ablation study.

\vspace{0.1cm}
\noindent $\bullet$ {\bf PQ-Net}~\cite{wu2020pq} learns a latent geometry representation with a Seq2Seq-based auto-encoder, followed by latent GAN~\cite{achlioptas2018learning} for code generation.
%generates 3D parts in two steps~\cite{wu2020pq}:
%is proposed to generate 3D parts in two major steps: 
%Seq2Seq-based auto-encoder and latent GAN for generating latent codes \cite{achlioptas2018learning}.
We keep the original auto-encoder architecture and modify the input/output as a sequence of roof primitives to fit our problem.
%In particular, the input and output is a sequence of roof primitives. 
Concretely, the input/output vector is a concatenation of boundary coordinates, facet angles, one-hot encoded primitive types, and the number of primitives. The training requires a pre-defined sequential order of primitives, for which we start from the largest rectangle, and then sort in a decreasing order of the distances to the largest one. 
%increasing distance to the largest one. 
We make no changes to the latent GAN.

\noindent $\bullet$ {\bf PQ-Net-Relation} is a novel variant of PQ-Net, where pairwise relationships are serialized into a one-hot vector with a fixed order and concatenated to the representation.
%further concatenated as one-hot vector by serialization with a fixed order.
%encoded relationships to the inputs and outputs of the auto-encoder. 
Note that the original PQ-Net did not encode relationships.
%paper of PQ-Net did not consider relationships.

\noindent $\bullet$ {\bf House-GAN} is a state-of-the-art floorplan generative model~\cite{nauata2020house}. We make two modifications: (1) %In addition to generating primitive masks for generator, 
At the end of the generator, we add two 4-layer CNNs (each followed by a linear layer) to output the facet angle vectors and the one-hot vector encoding the primitive types; (2) At the beginning of the discriminator, we add two linear layers to transform the facet angle vector and the primitive type vector to 512d, which is reshaped to $2\times16\times16$ and concatenated with the feature tensor of the primitive mask.

\noindent $\bullet$ {\bf \nickname~(w/o rela.)} is a variant of our architecture without the relationship generation. 
The relationship discriminator, the differentiable vectorization, and the differentiable relationship enforcement are also removed.
%: In the generator, we remove the branch of outputting relationships and generate the facet orientation/angle image only. 
%The other relationship-related modules including relationship discriminator, differentiable vectorization and differentiable relationship enforcement are all removed.

\noindent $\bullet$ {\bf \nickname~(w/o diff.)} is a variant without the differentiable vectorization and the differentiable enforcement modules during training.
These modules do not have learnable weights and are utilized during testing.
%The generator still outputs relationships and both discriminators are used. 

\vspace{0.1cm}

PQ-Net, House-GAN, and Roof-GAN (w/o rela.) do not generate relationship labels, and a simple snapping heuristic with a threshold is used. We vary the thresholds, but do not see much differences in the results. Therefore, we use an error tolerance of 1 pixel for colinearity and $18^\circ$ for parallelism, the same threshold setting used in the ground-truth preparation.
%All the above methods, except \nickname~(w/o diff.) 
%In the above methods, a simple thresholding is applied to check the colinearity/coplanarity relationships if the outputs do not contain relationship data. Here we use the same threshold setting with the one used in GT relationship creation.
%
House-GAN and Roof-GAN requires the relational graph, in particular, the number of primitives for a model generation. 
Both at training and testing time, we randomly pick the number of primitives by following the statistics of samples in our database, that is, ($30\%,51\%,14\%,5\%$) for houses with ($2,3,4,5$) primitives.
%
%When training House-GAN and Roof-GAN, the number of primitives has the same distribution with the GT samples. At test time, we randomly pick the primitive number from the set of $[2,5]$ with a uniform distribution.
PQ-Net and PQ-Net-Relation sequentially produces primitives until the termination probability reaches 0.5.
%outputs a probability of termination at each step, 
%, each time step outputs a probability depicting if the current step is the last primitive and the generation terminates if the probability is greater than 0.5.
We have 502 house samples before augmentation and randomly split them into  438 training and 64 testing samples. Each system generates 1,000 samples for the evaluation.

\begin{table}[!tb]
    \centering
    \caption{Quantitative evaluations. Three variants of the proposed approach (\nickname) are compared against the three competing methods on the two metrics. Recursive minimum matching distance (RMMD) is a new metric, while 
    %on the vector geometry representation in the unit of meter. 
    FID is a standard one.
    %, evaluating the realism and the diversity.
    %which evaluates the realism and diversity of 1,000 generated samples against 512 ground-truth samples, while taking into account the vector-structure of the generated roof models.
    %A standard FID metric is also calculated while a roof model is represented as a surface normal image. 
    %RMMD is calculated in the unit of meters. 
    The smaller the better for both metrics.
    %
    %comparison results using 64 test samples. 1000 samples are generated for each method. The unit of MMD is in meters, and the smaller the better for both metrics. 
    ``GAN'' column indicates if a method is GAN-based or not. ``Rela.'' column indicates if a method generates relationships or not, as opposed to threshood-based snapping.
    The colors \textcolor{cyan}{cyan} and \textcolor{blue}{blue} represent the best and the second best methods.}
    \vspace{0.2cm}
    \label{tab:quanti_results}   
    \begin{tabular}{l|c|c|c|c} \hline
    Method                    & GAN & Rela. & FID  & RMMD   \\ \hline
    PQ-Net                     & &    & 12.0 & 7.80\\
    PQ-Net-Relation            & &\checkmark    & 13.5 & 6.97\\
    House-GAN                  & \checkmark&    & 18.4 &7.43\\
    Roof-GAN (w/o rela.) & \checkmark & &  \textcolor{blue}{10.6} & 6.61\\
    Roof-GAN (w/o diff.)  & \checkmark &\checkmark &  11.9 &\textcolor{blue}{6.48}\\
    Roof-GAN              & \checkmark & \checkmark        &  \textcolor{cyan}{9.8} &  \textcolor{cyan}{6.20}\\ \hline
    \end{tabular}
\end{table}

\begin{table}[!tb]
    \centering
    \caption{Cross-validation results.
    To prevent a method from simply copying and pasting, we split the training and testing sets based on the number of primitives. See the text for the details.
    %Concretely, for the right columns (4 Rectangles) of the table, we train the networks on houses with 2, 3, or 5     
    %use the houses whose number of primitives are 2, 3, or 5
    %
    %The second and third columns are obtained by training on roofs with 2,4,5 rectangles and testing on roofs with 3 rectangles. The last two columns are obtained by training on roofs with 2,3,5 rectangles and testing on roofs with 4 rectangles.
    }
    \label{tab:cross_valid}
    \vspace{0.2cm}
    \begin{tabular}{l||c|c||c|c} \hline
    ~ & \multicolumn{2}{c||}{3 Primitives} & \multicolumn{2}{c}{4 Primitives} \\ \hline
    Method   & FID& RMMD   & FID & RMMD    \\ \hline
    PQ-Net   & 13.0 & 10.4 & 14.6 & 12.9  \\
    House-GAN & 27.5& 8.5  & 27.2 & 12.5  \\
    Roof-GAN & \textcolor{cyan}{11.1}& \textcolor{cyan}{7.5} & \textcolor{cyan}{13.8} & \textcolor{cyan}{10.9}  \\ \hline
    \end{tabular}
\end{table}

\begin{figure*}%[!p]
    \centering
    \includegraphics[width=0.98\textwidth]{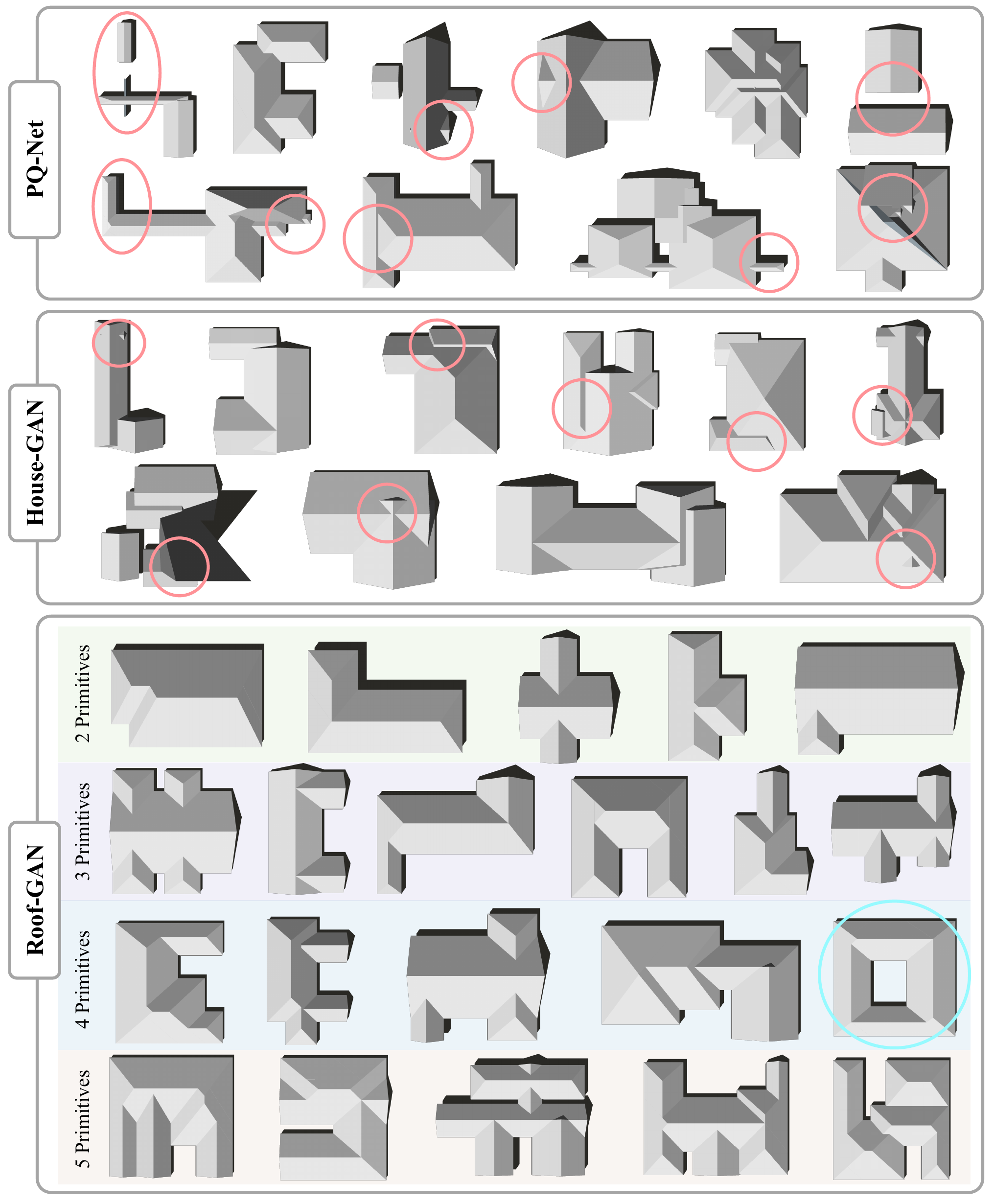}
    \caption{Qualitative evaluations among PQ-Net~\cite{wu2020pq}, House-GAN~\cite{nauata2020house}, and Roof-GAN (ours). \textcolor{red}{Red} ovals indicate non-realistic roof structures by the competing methods. Roof-GAN results are grouped row-by-row based on the number of primitives. Roof-GAN is capable of producing a new realistic roof structure that does not exist in our dataset, for example, an O-shaped building highlighted by the \textcolor{cyan}{cyan} oval.
    %comparisons of \nickname~to the two competing methods. \yasu{Maybe put red disk/ellipses where errors happen in (a) and (b)} \yasu{Put legend 2 primtives, 3 primitives for roof-gan at the side. Maybe have to use rectangle thing to show row-wise groupings better. Maybe simple repeat different colors row by row...}}
    }
    \label{fig:visual_result}
\end{figure*}

\begin{figure*}[!tb]
\centering
\includegraphics[width=0.95\textwidth]{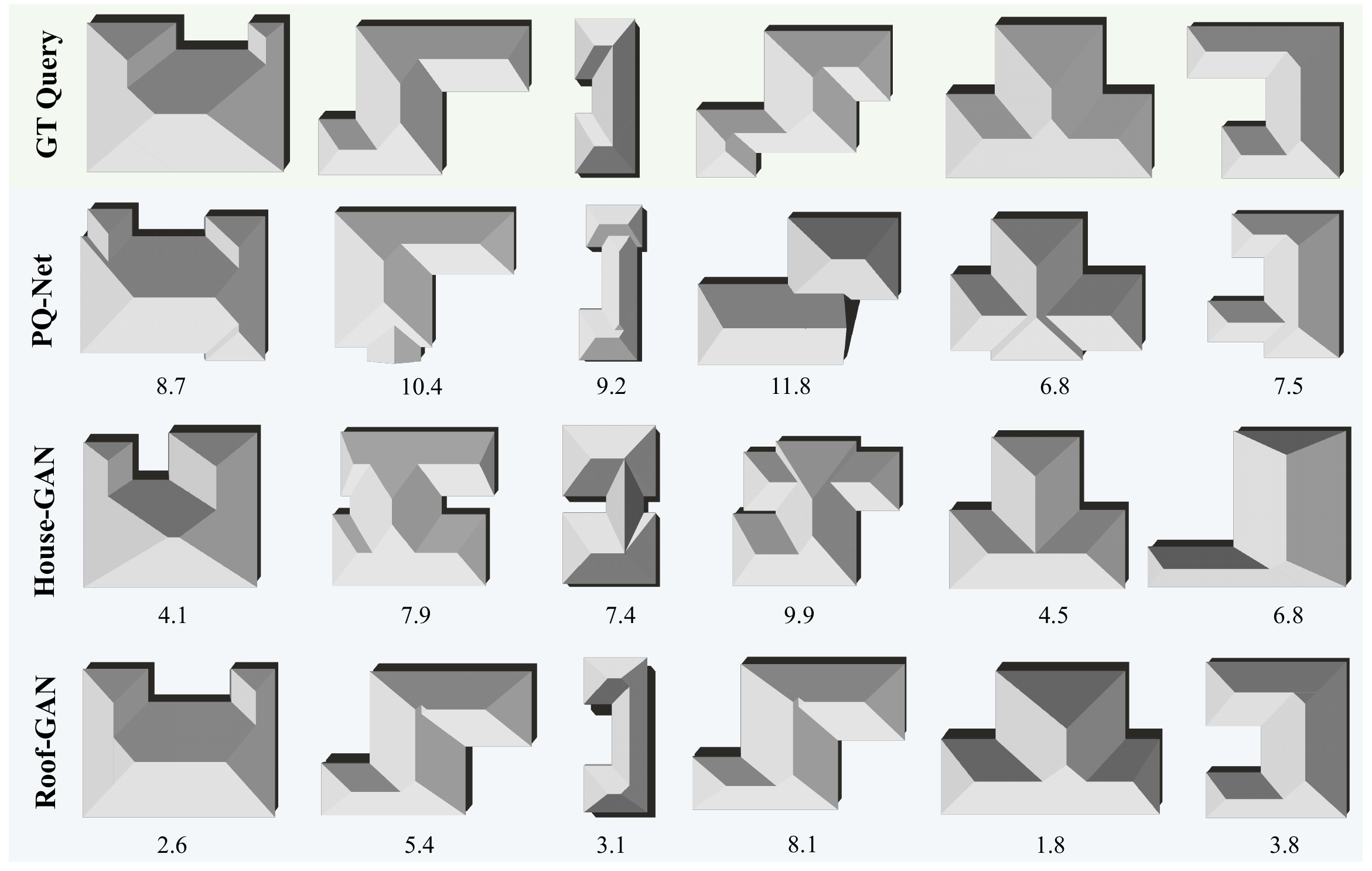}
\caption{RMMD-based model retrieval evaluations. 
In each column, given a query ground-truth model at the top, we use the RMMD-metric to find the closest sample among 1,000 generations by each method. Roof-GAN models are visually similar to the queries, which is also supported by the RMMD metric listed at the bottom of each model.
}
\label{fig:retrieval}
\end{figure*}

\subsection{Quantitative evaluations}

%We randomly split the dataset into training and testing. The test set contains 64 houses, and the others are for training. All the comparative methods and our \nickname~are trained on the same training set. 1000 samples are generated for each method and are compared to the test set.

%To eliminate the differences caused by rigid transformation, we apply data augmentation to each test set 

Table \ref{tab:quanti_results} shows the main quantitative evaluations, where
%results using the metrics of FID and MMD, where 
\nickname~makes clear improvements over all the competing methods on both metrics.
%and baselines in the aspects of both realism and diversity. 
PQ-Net is one of the state-of-the-art methods but has the largest error with the RMMD metric. PQ-Net-Relation is a novel baseline proposed in this paper, which encodes relationships in addition to the geometric parameters. PQ-Net-Relation reduces the error by roughly 11\%, which verifies the importance of relationship encoding, a key contribution of this paper.
%produces the largest MMD error. After including relationship data in auto-encoding, the error is greatly reduced, which demonstrates the importance of relationship reasoning for the problem roof structure generation. 
%
House-GAN is another existing state-of-the-art and better than PQ-Net by about 5\% in RMMD, but inferior to all our variants.
%
%in MMD metric but has larger FID error. This serves as a direct comparison between autoencoder-based method and GAN-based method. 

The last three rows of Table~\ref{tab:quanti_results} form the ablation study over the two technical contributions, relationship encoding and differentiable modules. 
Roof-GAN (w/o rela.) does not use relationship encoding and the differentiable modules, dropping the performance by about 6.6\% compared to Roof-GAN (all). The comparison between Roof-GAN (w/o rela.) and House-GAN sheds a light on our geometry representation  (Sect.~\ref{sec:representation}), which is in fact the only difference between the two systems. House-GAN uses vectorized representation for facet angles and primitive types while we use the new raster representation in Sect. \ref{sec:representation}.
Roof-GAN (w/o diff.) drops the performance about 4.5\%, validating the contributions of the differentiable modules.

The contributions of our innovations become even more significant in the cross validation experiments in Table~\ref{tab:cross_valid}. In order to prevent a method from copying and pasting models, we split the training and testing sets based on the number of primitives. Concretely, for the left columns (3 Primitives) of the table, we train the networks on houses with 2, 4, and 5 primitives and test on houses with 3 primitives. The performance gap further increases, where our RMMD score is better by 39\% than PQ-Net and by 13\% than House-GAN for the case of (3 Primitives).
%, verifying the power of the proposed architecture. 
Note that this evaluation may appear unfair for PQ-Net, which does not use the number of primitives in the test set. However, PQ-Net does not allow an external control
%A better way might be to limit the PQ-Net generations by the target primitive-count. However, PQ-Net
and rarely generate such samples (\ie, around 3\%), incapable of exploring new compositions.
%, where the quality is also poor.
%is not designed to explore new compositions, and  where the quality is also poor.
%with the target primitive count and such a scheme would be worse.

Roof-GAN outperforms all the other methods in the FID metric consistently. However, it is not clear why and how. For example, House-GAN is the worst in FID, despite reasonable RMMD scores.
RMMD has an intuitive physical meaning with a physical unit, and can also provide a distance measure for a particular pair of samples, which can be used for a model retrieval evaluation, all of which will be demonstrated in the next section.
%
%RMMD is another key contribution of the paper, which is a superior metric for structured geometry generation, capable of evaluating the realism and diversity like FID, while taking into account the vector structure of the geometry.

%When relationship-related modules are removed in our approach \ie, \nickname~(w/o relation output), the only difference between \nickname~and House-GAN is on the geometry representation, where House-GAN uses vectorized representation for facet angles and primitive types while we use the new structured representation introduced in Sect. \ref{sec:representation}. Our approach outperforms House-GAN in both MMD and FID at a large margin, validating the effectiveness of our new representation.

%The ablation study (the last three rows in Table \ref{tab:quanti_results}) shows that \nickname~is progressively improved as more relationship-related modules are included in the algorithm.

\subsection{Qualitative evaluations}
Figure~\ref{fig:visual_result} compares generated models by the three methods.
%PQ-Net, House-GAN, and \nickname.
Both PQ-Net and House-GAN generate unrealistic roof models. For example, PQ-Net produces isolated, too long, or too thin components.
Individual primitive shapes look much better in House-GAN, but poor incident relationships lead to unrealistic polygonal shapes and topology due to the failures of threshold based snapping.
%models with isolated blocks and models with long/thin rectangles. %
%Furthermore, due to the lack of relationship generation, nearby primitives or facets are not identified as colinear or coplanar by PQ-Net and House-GAN, generating some tiny spurs/facets and making the models look unrealistic. 
Roof-GAN, on the other hand, learns to generate incident relationships, producing complex and realistic combination of roof primitives.
%In comparison, \nickname~generates cleaner models, due to the colinearity and coplanarity relationship enforcement. 
The last four rows of the figure show the roof models with 2, 3, 4, and 5 primitives, respectively. 
%show the roofs with 2,3,4,5 rectangles from \nickname. 
Diversity of our generation is apparent in each row, especially in the last two rows. 
%in the generated samples is apparent
%One can clearly observe the great diversity in each row, especially for the last two rows. 
Compositional capability is another strength of Roof-GAN, which generates a new realistic roof structure such as an O-shaped building (highlighted by the \textcolor{cyan}{cyan} oval in Fig.~\ref{fig:visual_result}), which does not exist in our database.
%the last column second from the bottom

Lastly, Figure~\ref{fig:retrieval} demonstrates the model retrieval evaluations based on the RMMD metric. Each column shows a reference GT model at the top, and the closest sample by each method (from 1,000 generations). Roof-GAN is able to produce visually similar roof structures consistently, which is also supported by the RMMD metric.
%Notice that RMMD metric is capable of capturing intuitive distances between these models while respecting the underlying vector-structure. 
For example in the second column from the right, Roof-GAN is the only sample that has the same topology as the ground-truth. House-GAN sample is reasonable but misses a triangular facet in the middle and is penalized severely by the metric.
We provide more retrieval evaluations and the visualization of reconstructed models in the supplementary.

%\subsection{Generalization}
%To verify the generalization capability, we conduct cross-validation experiments. Specifically, to generate roofs with a certain number of primitives, we exclude samples with the same number of primitives from the GT data and use the remaining for training. For example, we train on roofs with 2,4,5 rectangles and test on roofs with 3 rectangles. Table \ref{tab:cross_valid} shows \nickname~performs the best in all metrics. We also observe that when training on roofs with 2,4,5 rectangles, only $3.3\%$ of the generated results of PQ-Net have 3 rectangles, meaning that PQ-Net tends to memorize the training data as an autoencoder-based approach.

%\ym{will also show visual results when generating roofs with more than 5 rectangles if we have space.}

\section{Conclusion}
This paper proposes a novel generative adversarial network that generates structured geometry of residential roof structures. 
For a roof represented as a graph, the generator learns to generate primitive geometries at the nodes and incident geometric relationships
%such as colinearity and coplanarity 
at the edges.
%the nodes representing the primitive raster-geometry and the edges representing inter-node colinearity and coplanarity.
%Differentiable vectorization and relationship enforcement modules convert the raster-geometry into a vector format. The raster-geometry, the vector-geometry, and the relationships are passed to the geometry and relationship discriminators in an end-to-end architecture.
%
%Two discriminators assess geometry and relationship respectively.
%
%We also develop novel differentiable modules which convert raster-geometory to vector-geometry while enforcing the generated relationships. 
Qualitative and quantitative evaluations demonstrate the effectiveness of our approach in generating diverse and realistic roof models over all the competing methods.
%, where we propose a novel metric for the task of structured geometry generation.
%that measures the realism and diversity of generated samples while taking 
Our future work includes the handling of higher-order primitive relationships such as symmetries and more diverse and complex primitive types such as dormers and chimneys. 
%We will share our code and data.
%In the future, we plan to consider more relationship types such as layout symmetry across primitives and generate more sophisticated roof elements such as dormer and chimney. We will release our code and data.
%\input{planning.tex}

\vspace{2pt}
\mysubsubsubsection{Acknowledgements} This research is partially supported
by NSERC Discovery Grants (No. 611714 and No. 611370), an NSERC Discovery Accelerator Supplement, and an DND/NSERC Discovery Grant
Supplement.

\clearpage
%\nextpage

{\small
\bibliographystyle{ieee_fullname}
\bibliography{egbib}
}

\end{document}